\documentclass[runningheads]{llncs}
\usepackage{graphicx}
\usepackage[noline,ruled]{algorithm2e}
\usepackage{tikz}
\usepackage{comment}
\usepackage{array}
\usepackage{amsmath,amssymb} 
\usepackage{color}
\usepackage{bm}
\usepackage{multirow}
\usepackage{wrapfig,lipsum}

\newif\ifmodify
\definecolor{LightCyan}{rgb}{0.88,1,1}

\ifmodify
\newcommand{\ziquan}[1]{\textcolor{orange}{#1}}
\newcommand{\cross}[1]{\textcolor{red}{\sout{#1}}}
\newcommand{\abc}[1]{\textcolor{blue}{#1}}

\else
\newcommand{\cross}[1]{}
\newcommand{\ziquan}[1]{#1}
\newcommand{\abc}[1]{#1}

\fi

\newcommand{\CUT}[1]{}

\title{Boosting Adversarial Robustness From The Perspective of Effective Margin Regularization}

\titlerunning{Boosting Adversarial Robustness From The Perspective of EMR}
\author{Ziquan Liu\inst{1} \and
Antoni B. Chan\inst{1}}
\authorrunning{Liu and Chan}
\institute{Department of Computer Science, City University of Hong Kong\\
\email{ziquanliu2-c@my.cityu.edu.hk, abchan@cityu.edu.hk}}

\newcommand{\loss}{\mathcal{L}}
\newcommand{\btheta}{\bm{\theta}}
\newcommand{\bW}{\bm{W}}
\newcommand{\bw}{\bm{w}}
\newcommand{\bp}{\bm{p}}

\newcommand{\bbias}{\bm{b}}
\newcommand{\bx}{\bm{x}}
\newcommand{\bdelta}{\bm{\delta}}

\newcommand{\bl}{\bm{l}}
\newcommand{\real}{\mathbb{R}}
\newcommand{\datadis}{\mathcal{D}}
\newcommand{\alg}{\mathcal{A}}

\DeclareMathOperator*{\argmax}{arg\,max}

\begin{document}

\maketitle

\begin{abstract}
The adversarial vulnerability of deep neural networks (DNNs) has been actively investigated in the past several years. 
This paper investigates the scale-variant property of cross-entropy loss, which is the most commonly used loss function in classification tasks, and its impact on the effective margin and adversarial robustness of deep neural networks. 
Since the loss function is not invariant to logit scaling, increasing the effective weight norm will make the loss approach zero and its gradient vanish while the effective margin is not adequately maximized. On typical DNNs, we demonstrate that, if not properly regularized, the standard training does not learn large effective margins and leads to adversarial vulnerability.  To maximize the effective margins and learn a robust DNN, we propose to regularize the effective weight norm during training. Our empirical study on feedforward DNNs demonstrates that the proposed effective margin regularization (EMR) learns large effective margins and boosts the adversarial robustness in both standard and adversarial training. On large-scale models, we show that EMR outperforms basic adversarial training, TRADES and two regularization baselines with substantial improvement. Moreover, when combined with several strong adversarial defense methods (MART~\cite{wang2019improving} and MAIL~\cite{liu2021probabilistic}), our EMR further boosts the robustness. 
\end{abstract}

\section{Introduction}
One major challenge to the security of computer vision systems is that deep neural networks (DNNs) often fail to achieve a satisfactory performance under adversarial attacks \cite{szegedy2013intriguing}. Since the phenomenon is observed, various adversarial attacks \cite{goodfellow2014explaining,carlini2017towards,croce2020reliable} and defense methods \cite{kurakin2016adversarial,madry2017towards,zhang2019theoretically} have been proposed and the understanding into the adversarial vulnerability of DNNs is improved \cite{bachmann2021uniform,tsipras2018robustness,ilyas2019adversarial}. Denote the DNN 
as $f_{\btheta}\colon \bx \mapsto \bl$, with $\bx \in \real^{D}$ and $\bl \in \real^{K}$. The model is optimized by algorithm $\alg$ that minimizes empirical risk $\loss$ over training set $\datadis_{tr}$,
\begin{align}
\btheta^* = \alg(f_{\btheta},\datadis_{tr},\loss).
\end{align}
There are generally four \emph{direct} methods to improve the robustness of DNNs. First, the function space $f_{\btheta}$ can be designed to accommodate the need for adversarial robustness. For example, replacing the piecewise linear activation with a smooth activation function improves the performance of adversarial training \cite{xie2020smooth},  and some architectural configurations are better than others in terms of adversarial robustness \cite{huang2021exploring}. Second, the algorithm $\alg$ can be incorporated with inductive biases to learn a function with some specific properties, such as low model complexity \cite{krogh1991simple}, local linearization \cite{qin2019adversarial} and feature alignment \cite{li2021towards}. Third, the training set $\datadis_{tr}$ can be shifted by adversarial perturbation \cite{madry2017towards,zhang2019theoretically} or other data augmentation \cite{gowal2021improving,rebuffi2021data} to enhance the robustness. Finally, a carefully designed loss $\loss$ can be used to improve the robustness, such as Max Mahalanobis center loss \cite{pang2019rethinking}. Besides the direct adversarial defenses, \emph{indirect} adversarial defenses are also investigated, e.g., adversarial examples detection \cite{pang2018towards,lu2017safetynet,xu2017feature,roth2019odds} and obfuscated gradient defenses \cite{athalye2018obfuscated,ma2018characterizing,buckman2018thermometer,dhillon2018stochastic,guo2018countering,xie2017mitigating,song2017pixeldefend,samangouei2018defense}. 

\begin{figure}[t]
    \centering
    \includegraphics[width=1.0\textwidth]{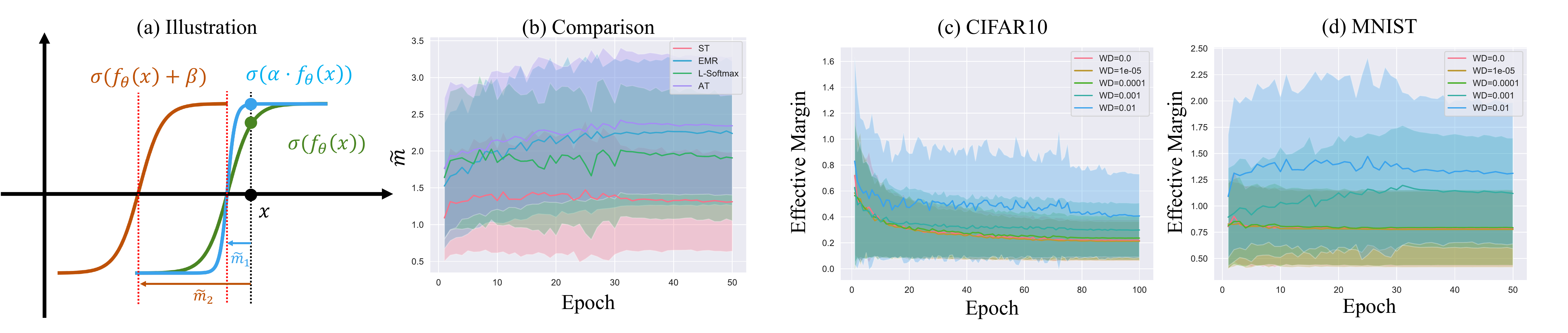}
    
    \caption{The problem of cross-entropy loss in maximizing effective margins and the proposed ERM's performance in terms of increasing the effective margins on MNIST test set. \textbf{(a)} The gradient of the sigmoid function $\sigma(\cdot)$ vanishes if we scale up the input logit $f_{\btheta}(\bx)$ by $\alpha>1$. However, its distance to the decision boundary (dashed red vertical line), i.e., the effective margin $\tilde m$ defined in (\ref{eqn:eff_margin}) remains the same. This property shows that only training with cross-entropy loss does not effectively increase the actual margin. Our work aims to push the decision boundary away from the sample so that the $\tilde m$ is increased, e.g., $f_{\btheta}(\bx)+\beta$ with $\beta>0$.
    \textbf{(b)} The means (solid lines) and standard deviations (background shadow) of the effective margins of an MLP on the MNIST test set. The proposed EMR achieves an $\tilde m$ and adversarial robustness that is comparable with adversarial training,  and outperforms standard training with weight decay or L-Softmax.  See Table \ref{tab:mnist_mlp} for details. \textbf{(c)} and \textbf{(d)}  Effective margin on CIFAR10 and MNIST when different $\lambda_{WD}$ are used. 
    Training without weight decay (WD) or with small $\lambda_{WD}$ leads to smaller effective margins.
   }
    \label{fig:illustrative}
\end{figure}
This work falls into the second category \abc{(inductive bias)} where the neural network is trained with regularization to boost the adversarial robustness. We consider the most popular loss function for the classification task, the cross-entropy (XE) loss,
\begin{align}
XE_i = -\sum_{k=1}^K y_{ik}\log\frac{\exp(l_{ik})}{\sum_j \exp(l_{ij})},
\end{align}
where the logit $l_{ik}=f_{\btheta}^{(k)}(\bx_i)$ is the $k$-th output of the neural network \abc{for the $i$-th sample}. One property of the network is that the prediction for the $i$-th sample, i.e., $\hat y_i=\argmax_{k\in[K]} l_{ik}$, is invariant to scaling the \ziquan{logit vector} $\bl_i=[l_{ik}]_k$  by a positive constant $\alpha$. In other words, the classification accuracy will not change if we scale $\bl_i$ up to $\alpha\bl_i$ where $\alpha>1$. 
However, the XE loss will vanish if we scale up the logit. 

This phenomenon brings a problem in optimization with XE loss, since the training only aims to minimize the loss without maximizing the \emph{effective} margin, which is defined as the normalized logit difference (see Equation \ref{eqn:eff_margin}) and is invariant to the weight magnitudes. Once a sample is correctly classified, the scale-variant property of XE loss can be exploited by SGD to minimize the loss while the distance to the decision boundary  \abc{(in the input space)}  remains small (see Fig.~\ref{fig:illustrative}a). In a homogeneous NN \cite{lyu2019gradient,du2018algorithmic,liu2020improve}, such as multi-layer perceptron (MLP) and convolutional neural network (CNN) without residual connections or normalization layers, the logit magnitude scales with weight norms so the training algorithm can minimize the loss by increasing weight norms. In a ResNet \cite{he2016deep}, the final classification layer can be scaled up to minimize the cross-entropy loss. Weight decay \cite{krogh1991simple} is a common strategy to increase the effective margin by controlling the squared $l_2$ norm of weights in the DNN. However, it is known that adversarial robustness cannot be achieved by only using weight decay (WD), especially in deeper networks \cite{szegedy2013intriguing}. The most popular way to robustify DNNs is adversarial training (AT) \cite{madry2017towards}, which \emph{explicitly} perturbs samples to be on the desired margin from the original samples, and then trains on the perturbed samples.
However, AT incurs increased computational cost for training due to the generation of the adversarial training samples in each iteration.

In this paper, we propose \emph{effective margin regularization} (EMR) to push the decision boundary away from the samples 
by controlling the effective weight norms of the samples. We first show that traditional regularization such as weight decay and large-margin loss (e.g., \cite{liu2016large}) cannot train a DNN with satisfactory robustness. Then the proposed method is compared with WD, large-margin softmax and adversarial training,  where we show its strength at maximizing the effective margin and thus improving adversarial robustness. Finally, on large-scale DNNs, 
we propose an approximation to EMR and demonstrate that when combined with adversarial training, EMR achieves competitive results compared with basic adversarial training and two recent regularization methods for improving adversarial training, i.e., Input Gradient Regularization (IGR) \cite{ross2018improving} and Hypersphere Embedding (HE) \cite{pang2020boosting}. Note that our EMR is complementary to adversarial training (AT) -- EMR pushes the decision boundary away from the training samples so as to increase the effective margin, while AT generates training samples on the desired margin. Thus EMR and AT can be combined to further improve adversarial robustness.


\section{Related Work}
\noindent\textbf{Adversarial Defense.} The standard way to train an adversarially robust DNN is to use adversarial training \cite{madry2017towards}. The clean examples are deliberately perturbed to approach the desired margin distance, so that the effective margin is produced during training.
Based on adversarial training, regularization approaches are proposed to learn a DNN 
with desired properties. \cite{ross2018improving} proposes to regularize the norm of the loss gradient with respect to input (IGR). In contrast, our work proposes to regularize the gradient of \emph{logit} with respect to input to maximize the effective margin. Locally Linear Regularization (LLR) \cite{qin2019adversarial} is proposed to learn a more linear loss function at each training sample, while our paper controls the local logit function's weight norm for training samples. Hypersphere embedding (HE) \cite{pang2020boosting} proposes to normalize the features and classification layer to alleviate the influence of weight norms. In our empirical study, we demonstrate that EMR achieves better robustness than IGR and HE on large-scale neural networks.

\noindent\textbf{Margin Regularization.} The hinge loss \cite{crammer2001algorithmic} is a classical loss to induce a 
large margin in SVM \cite{crammer2001algorithmic}. On DNNs, several losses are proposed to induce large margins, such as 
 Large-Margin Softmax \cite{liu2016large}, A-Softmax \cite{liu2017sphereface} and AM-Softmax
\cite{wang2018additive}. These large-margin losses still have problems to learn large effective margins since the scale of features and weights affects the loss values. On both MLP and CNN, we demonstrate that training with L-Softmax loss improves the effective margin compared with the standard cross-entropy loss, while EMR learns a larger effective margin than L-Softmax since EMR considers the scale problem in the loss function. \cite{lyu2019gradient,chizat2020implicit} study the normalized margin of homogeneous DNNs 
trained with gradient descent from a theoretical perspective and prove that the normalized margin is maximized by the gradient descent. Our work empirically investigates the normalized margin in DNNs trained with \emph{stochastic} gradient descent and its influence on the adversarial robustness. We show that by controlling the effective weight norm and increasing the effective margin, the adversarial robustness can be improved over vanilla training with SGD and WD. The attack method DeepFool \cite{moosavi2016deepfool} moves an input sample to cross its decision boundary by treating the model as a linear classifier at each optimization step, which is related to the margin of a classifier. In contrast, our paper proposes to defend against adversarial attacks by increase the effective margin during training. We did not evaluate the DeepFool since it is not a standard attack method in adversarial defense literature \cite{croce2020robustbench} and our experiment shows that DeepFool is not as effective as PGD at attacking large-scale models. Max-Margin Adversarial training (MMA) \cite{Ding2020MaxMarginA} proposes to approximate the margin by \ziquan{pushing input samples to cross the \abc{classification} boundary with PGD and recording the moved distance}.
 In contrast, EMR proposed to maximize the effective margin by regularizing the effective weight matrix norm, and can boost the adversarial robustness of both standard training and adversarial training. Since the performance of MMA is worse than vanilla PGD and TRADES according to AutoAttack benchmark \cite{croce2020reliable}, we do not include the comparison with MMA in the experiment. 

\section{Regularizing Effective Weight Norm Improves Effective Margin and Adversarial Robustness}
\CUT{
\begin{figure}[t]
    \centering
    \includegraphics[width=1.0\textwidth]{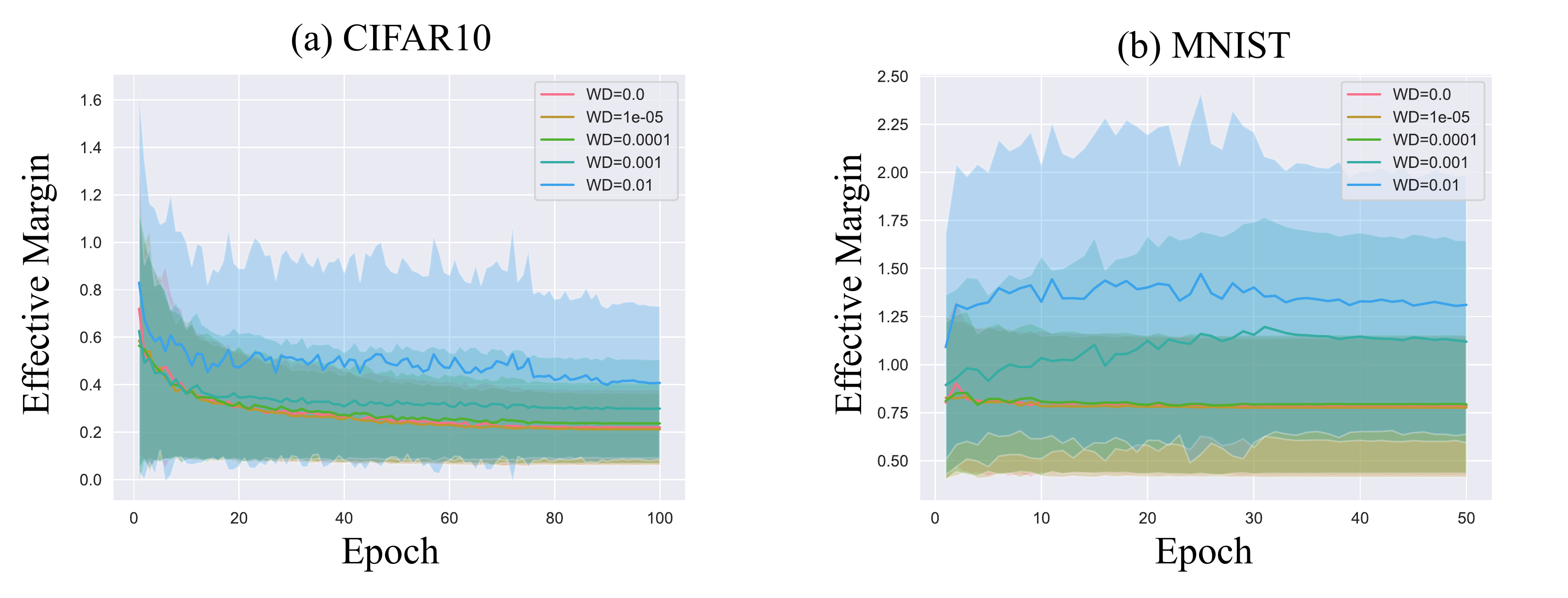}
    \vspace{-0.8cm}
    \caption{Effective margin on CIFAR10 and MNIST when different $\lambda_{WD}$ are used. 
    Training without weight decay (WD) or with small $\lambda_{WD}$ leads to smaller effective margins.}
    \vspace{-0.4cm}
    \label{fig:WD_eff_margin}
\end{figure}
}

\begin{figure}[t]
    \centering
    \includegraphics[width=1.0\textwidth]{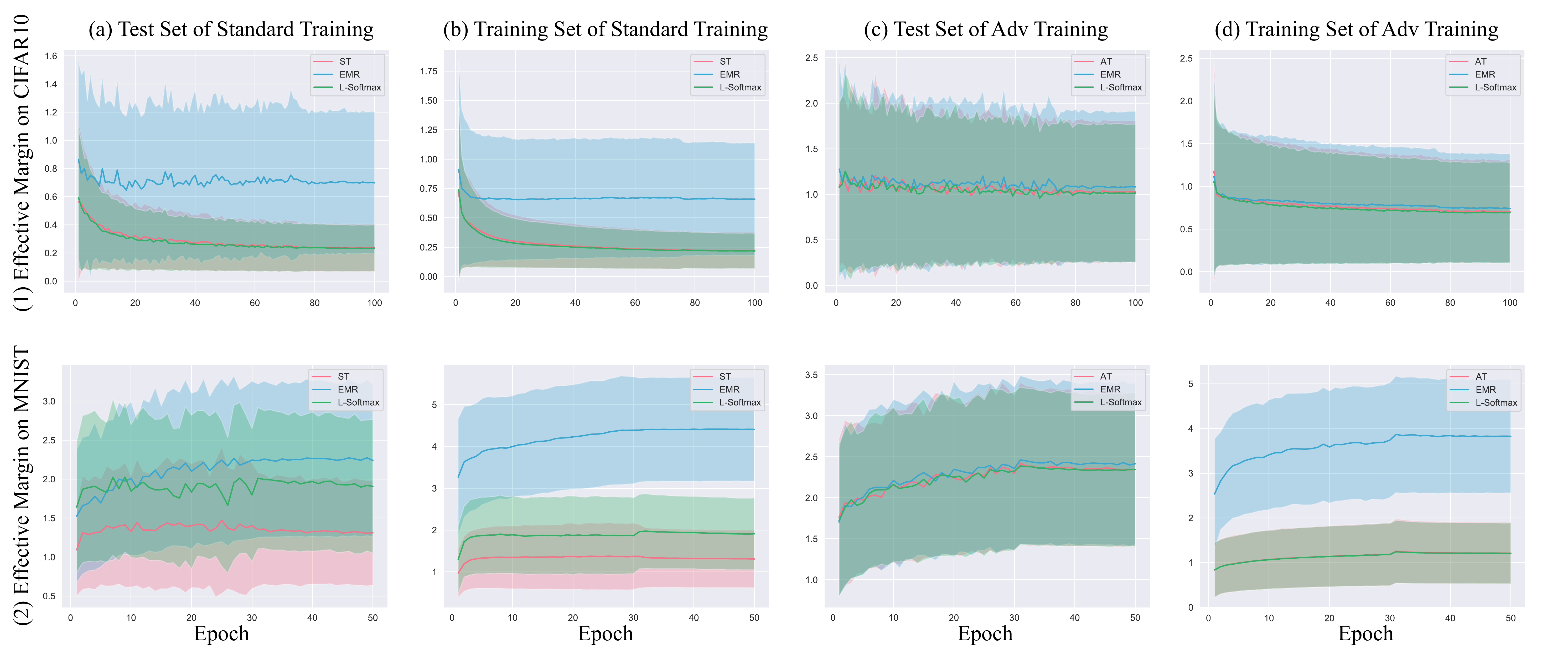}
    \caption{Effective margins on the training and test sets of CIFAR10 and MNIST. In normal training, EMR has a significant advantage over L-Softmax and weight decay (WD). In adversarial training, the improvement of EMR is still observable. 
    }
    \label{fig:eff_margin}
\end{figure}

\CUT{
\begin{figure}[t]
    \centering
    \includegraphics[width=1.0\textwidth]{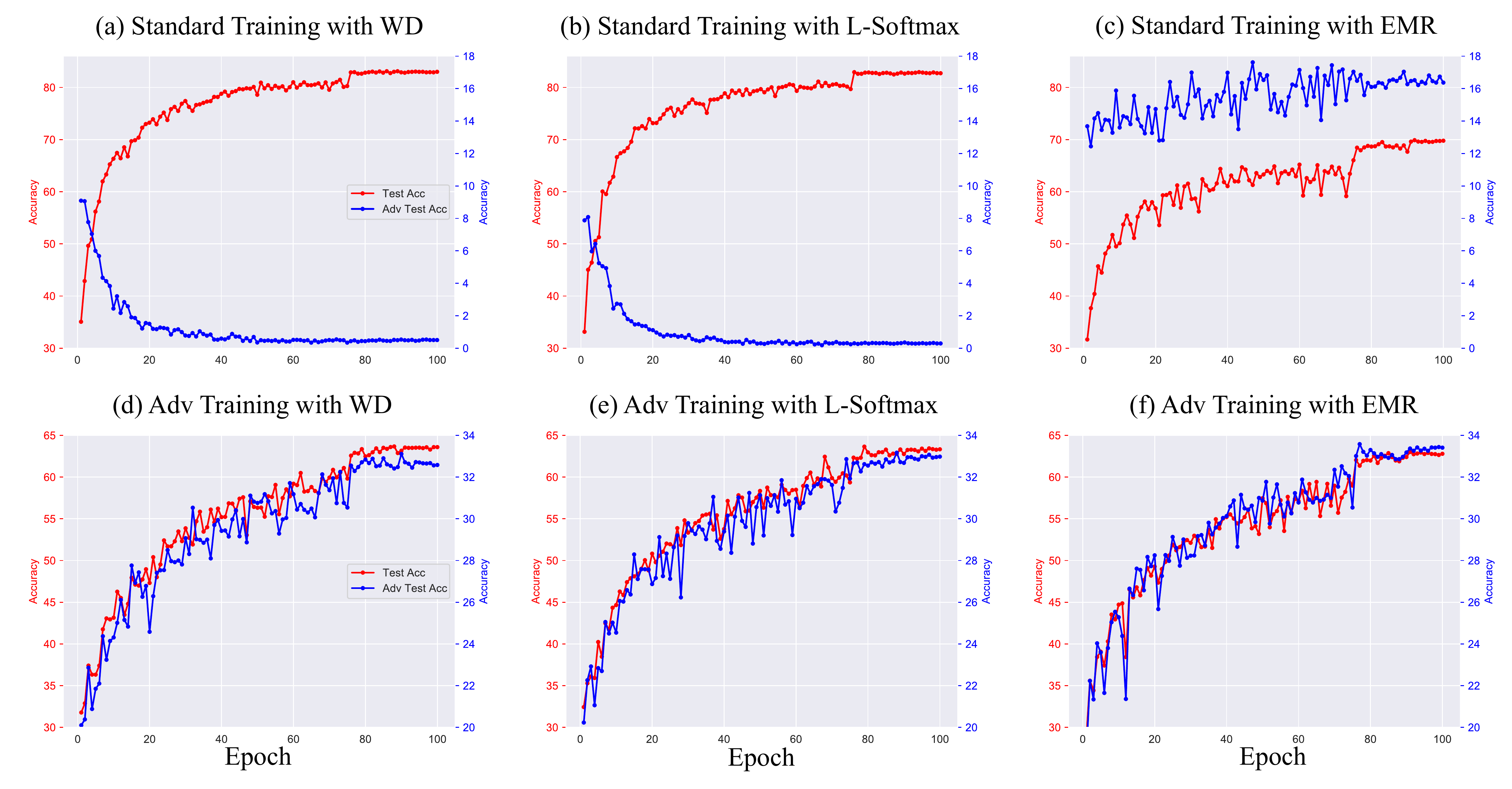}
    \vspace{-0.8cm}
    \caption{Training curve on CIFAR10 with 4-hidden layer CNN. In standard training, EMR has 
    a higher robust test accuracy than training with WD or L-Softmax. In adversarial training, EMR also achieves the best robust accuracy.
    }
    \vspace{-0.6cm}
    \label{fig:cifar_train_curve}
\end{figure}
}

\CUT{
\begin{figure}[t]
    \centering
    \includegraphics[width=1.0\textwidth]{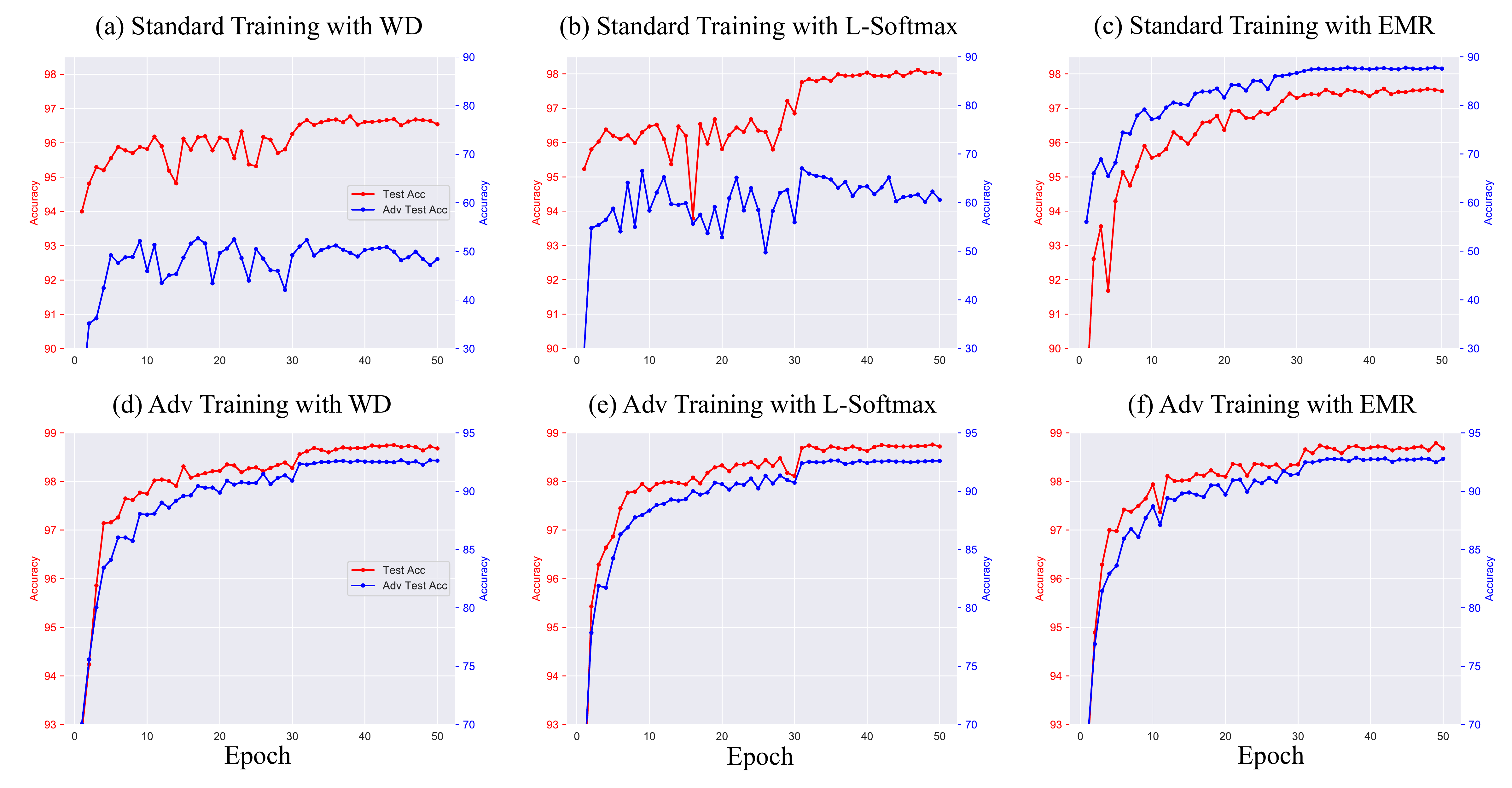}
    \vspace{-0.8cm}
    \caption{Training curve on MNIST with 4-hidden layer MLP. In normal training, EMR achieves significantly higher robust test accuracy than two baselines and is even comparable with that of adversarial training.}
    \vspace{-0.5cm}
    \label{fig:mnist_train_curve}
\end{figure}
}

We use the notation for a DNN in Section 1. A general DNN, such as MLP, CNN and evaluation-mode Resnet with piece-wise linear activation functions (e.g., ReLU and LeakyReLU), can be expressed as a linear function for \emph{each} input sample, i.e., $f_{\btheta}(\bx_i) = \bW(\bx_i)\bx_i + \bbias(\bx_i)$. The weight matrix \abc{$\bW_i\triangleq\bW(\bx_i)$} and bias \abc{$\bbias_i\triangleq\bbias(\bx_i)$} are determined by input samples since the activations will change with the input.
Similar to the normalized margin in \cite{lyu2019gradient}, we define the \emph{effective margin} $\tilde m_i$ as the normalized logit difference between the ground-truth class and the closest other class,
\begin{align}
    \tilde m_i = \min_{j\neq y_i} \tfrac{f_{\btheta}^{(y_i)}(\bx_i)-f_{\btheta}^{(j)}(\bx_i)}{\|\bw_i^{(y_i)}-\bw_i^{(j)}\|_2},
    \label{eqn:eff_margin}
\end{align}
where $y_i$ is the ground-truth class for $\bx_i$ and $\bw_i^{(j)}$ is the $j$th row vector of $\bW_i$. 
The quantity is relevant to adversarial robustness, since it describes the actual distance of a sample to the decision boundary in the input space. Thus, to improve adversarial robustness, it is desirable to maximize the effective margin, \abc{so that adversarial examples will fall inside the margin but still correctly classified}. However, during training a DNN, once a training image is correctly classified by the network, the scale of $f_{\btheta}(\bx_i)$ can be increased to minimize the loss without effectively maximizing the actual margin. We aim to alleviate this problem by regularizing the effective weight norm during training.

For simplicity, we first consider a DNN without residual connections \ziquan{and assume that the bias term is appended as an additional dimension of the weight and $\bx_i:=[\bx_i;1]$; residual DNNs are considered in the experiment section.}
 \abc{The logit output of the network is determined by the angle $\phi_{ij}$ between $\bx_i$ and $\bw_i^{(j)}$, and the lengths $\|\bx_i\|$ and $\|\bw_i^{(j)}\|$,}
\begin{align}
    l_{ij} = \abc{\bx_i^T \bw_i^{(j)}} = \|\bx_i\|\|\bw_i^{(j)}\|\cos(\phi_{ij}).
\end{align}
\begin{table}[t]
    \centering
    \newcolumntype{S}{>{\centering\arraybackslash} m{1.3cm}}
    \newcolumntype{M}{>{\centering\arraybackslash} m{0.5cm}}
    \begin{tabular}{c|c|c|c|c|c}
         Training & $\lambda_{WD}$ & Clean Acc. & PGD20 & $\tilde m_{\text{train}}$ & $\tilde m_{\text{test}}$  \\
         \hline
        ST &0.1 & 77.88 & 34.06  & 1.00$\pm$0.69 & 1.02$\pm$0.70 \\
        ST &0.01 & 96.54 & 48.41  & 1.31$\pm$0.69 & 1.31$\pm$0.67\\
        ST & 0.001& 98.41 & 24.41  & 1.11$\pm$0.51 & 1.12$\pm$0.53\\
        ST &0.0001 & 98.33 & 4.69  & 0.79$\pm$0.34 & 0.79$\pm$0.36\\
        ST+LSoftmax &0.1& 17.83 & 17.27  & 3.13$\pm$1.11 & 3.21$\pm$1.16\\
        ST+LSoftmax &0.01 & 98.00 & 60.59  & 1.91$\pm$0.86 & 1.91$\pm$0.85\\
        ST+LSoftmax & 0.001& \textbf{98.63} & 51.07  & 1.82$\pm$0.63 & 1.81$\pm$0.66\\
        ST+LSoftmax & 0.0001 & 98.50 & 28.91  & 1.34$\pm$0.44 & 1.34$\pm$0.47\\
        ST+EMR$_{0.1}$ & 0.001 & 97.50 & \textbf{87.56} & 4.41$\pm$1.24  & 2.24$\pm$0.98\\
        \hline
        AT &0.01 & 97.66 & 90.11 & 1.18$\pm$0.69 & 2.27$\pm$1.07 \\
        AT &0.001 & 98.68 & 92.62 & 1.21$\pm$0.69 & 2.34$\pm$0.94 \\
        AT &0.0001 & \textbf{98.98} & 92.24 & 1.09$\pm$0.61 & 2.13$\pm$0.77 \\
        AT+LSoftmax &0.01& 97.57 & 89.88 & 1.18$\pm$0.69 & 2.28$\pm$1.07 \\
        AT+LSoftmax &0.001& 98.72 & 92.60 & 1.21$\pm$0.68 & 2.35$\pm$0.94 \\
        AT+LSoftmax &0.0001& 99.02 & 92.50 & 1.10$\pm$0.63 & 2.12$\pm$0.78 \\
        AT+EMR$_{3e-4}$  & 0.001 & 98.68 & \textbf{92.78} & 3.83$\pm$1.27  & 2.42$\pm$0.99\\
        \hline
    \end{tabular}
    \caption{Adversarial robustness and effective margins of MLP on MNIST  for standard training (ST) and adversarial training (AT). PGD20 attack has an $l_{\infty}$ bound of $\epsilon=0.1$ and the step size $\alpha$ is 0.01. The mean and standard deviation of effective margins defined in Equation \ref{eqn:eff_margin} of training ($\tilde m_{\text{train}}$) and test ($\tilde m_{\text{test}}$) sets are shown.
       }
    \label{tab:mnist_mlp}
\end{table}

The range of input magnitudes $ \|\bx_i\|$ is often fixed in the training/test stages, e.g., normalizing pixel values to $[0.0,1.0]$.  Thus, in order to minimize the loss, the training aims to increase $l_{iy}$ and decrease $l_{ij},\forall j\neq y$, by updating $\|\bw_i^{(j)}\|$ and $\cos(\phi_{ij})$. As we want to learn a large effective margin, it is beneficial to constrain the weight norm $\|\bw_i^{(j)}\|$ during training and let the optimization focus on the angular distance. Weight decay is a common method to control the weight norm \abc{of individual layers}, while other works have proposed large margin losses \cite{liu2016large}. In contrast, here we propose effective margin regularization (EMR) to directly penalize the \emph{effective} weight norm for training samples, i.e.,
\begin{align}
    \loss = \frac{1}{B}\sum_i^BXE(f_{\btheta}(\bx_i),y_i) +\lambda_{EMR}\frac{1}{B}\sum_i^B\sum_j^K\|\bw_i^{(j)}\|^2.
    \label{eqn:emr}
\end{align}
Different from WD, which regularizes all parameters of a DNN, our EMR regularizes the \emph{local} weight norms of the training samples. In our implementation, to compute the effective weight matrix for each output dimension, we loop over the object categories and take the gradient of the sum of  the $j$-th logit over a batch, i.e., $\sum_il_{ij}$, with respect to the input batch, and obtain $\bw_i^{(j)}, \forall i$ as a result of computational independence of samples. The $\lambda_{WD}$ is the coefficient for the WD regularization, where the squared L2 norm of all parameters is added to the loss. In contrast, we regularize a weighted sum of L2 norm of the logit’s gradient (effective weight norm) as in Equation \ref{eqn:emr}.

\begin{table}[t]
    \centering
    \newcolumntype{S}{>{\centering\arraybackslash} m{1.3cm}}
    \newcolumntype{M}{>{\centering\arraybackslash} m{0.5cm}}
    \begin{tabular}{c|c|c|c|c|c}
         Training & $\lambda_{WD}$ & Clean Acc. & PGD10 & $\tilde m_{\text{train}}$ &  $\tilde m_{\text{test}}$\\
         \hline
        ST&0.01& 55.77 & 4.19 & 0.39$\pm$0.31 & 0.41$\pm$0.32 \\
        ST & 0.001 & 82.10 & 0.68 & 0.28$\pm$0.19 & 0.30$\pm$0.21 \\
        ST &0.0001& \textbf{83.00} & 0.50 & 0.22$\pm$0.15 & 0.23$\pm$0.17 \\
        ST+LSoftmax &0.01 & 55.06 & 4.90 & 0.41$\pm$0.33 & 0.43$\pm$0.34 \\
        ST+LSoftmax &0.001& 82.39 & 0.81 & 0.27$\pm$0.18 & 0.28$\pm$0.19 \\
        ST+LSoftmax &0.0001& 82.73 & 0.29 & 0.22$\pm$0.15 & 0.23$\pm$0.16 \\
        ST+EMR$_{0.01}$ &0.0005 & 69.78 & \textbf{16.37} & 0.66$\pm$0.48 & 0.70$\pm$0.50\\
        \hline
        AT &0.01 & 34.73 & 22.98 & 0.90$\pm$0.82 & 1.22$\pm$1.04 \\
        AT &0.001& 60.27 & 32.47 & 0.76$\pm$0.65 & 1.09$\pm$0.84 \\
        AT &0.0001& \textbf{63.59} & 32.58 & 0.71$\pm$0.60 & 1.03$\pm$0.78 \\
        AT+LSoftmax &0.01 & 34.62 & 22.87 & 0.91$\pm$0.82 & 1.21$\pm$1.02 \\
        AT+LSoftmax &0.001& 60.47 & 32.29 & 0.76$\pm$0.65 & 1.08$\pm$0.84 \\
        AT+LSoftmax &0.0001& 63.09 & 32.96 & 0.71$\pm$0.61 & 1.03$\pm$0.78 \\
        AT+EMR$_{0.001}$&0.0005& 62.79 & \textbf{33.41}  & 0.74$\pm$0.64 & 1.08$\pm$0.83\\
        \hline
    \end{tabular}
    \caption{Adversarial robustness and effective margins of CNNs on CIFAR10. PGD10 attack has an $l_{\infty}$ bound of $\epsilon=0.031$ and the step size $\alpha$ is 0.0078. 
    }
    \label{tab:cifar_cnn}
\end{table}

\noindent\textbf{Empirical Results for Standard Training.} We test the performance of EMR on two common feedforward neural networks, MLP and CNN. The MLP consists of 4 hidden layers and one output layer with the hidden dimensions as 1,024. The model is trained using SGD+Momentum on MNIST \cite{lecun1998mnist} for 50 epochs and with a batch size of 100. The initial learning rate is 0.01 and we divide it by 10.0 at the 30th epoch. The CNN consists of 4 convolution layers and the detailed architecture is in supplemental. The model is trained using SGD+Momentum on CIFAR10 \cite{krizhevsky2009learning} for 100 epochs and with a batch size of 100. The initial learning rate is 0.01 and we divide it by 10.0 at 75th and 90th epochs. 
We compare standard WD, L-Softmax and EMR on the CNN and MLP. On MLP, the margin parameter of L-Softmax loss is set as 4, while on CNN, the margin parameter is 1, otherwise the training will fail to converge. 
To evaluate the robustness, PGD attack \cite{madry2017towards} with $l_{\infty}$ norm bound is used. For CNN,  we use PGD10 with step size $\alpha=0.0078$ and norm bound $\epsilon=0.031$. For MLP, we use PGD20 with step size $\alpha=0.01$ and norm bound $\epsilon=0.1$. 

In Fig.~\ref{fig:illustrative}c-\ref{fig:illustrative}d, we first plot the effective margin of test samples during model training when standard WD is used with different hyperparameters $\lambda_{WD}$.
Note that we only plot $\tilde m_i$ of correctly classified images since those images are the targets of adversarial attacks. The effective margin when WD is not used is clearly smaller than imposing a large weight decay, which validates our argument that penalizing large weight norms helps increase the effective margin in XE loss optimization. 
Fig.~\ref{fig:eff_margin}a-b show the effective margin of training and test samples for standard training (ST).  L-Softmax has a benefit on the effective margins for MLP, but not CNN. In contrast, on both architectures, EMR achieves the highest effective margin. We show the clean and robust test accuracy curves during CNN and MLP training in the supplemental. 

Table \ref{tab:mnist_mlp} (top) shows the evaluation results of MLP when using standard training  with three methods. Two key observations are that: a) increasing $\lambda_{WD}$ cannot achieve adversarial robustness that is comparable with L-Softmax or EMR, indicating that more advanced approaches are needed to maximize the effective margin; b) the adversarial robustness of training with EMR is substantially higher than training with L-Softmax, demonstrating the importance of regularizing effective weight norms. Table \ref{tab:cifar_cnn} (top) shows the robust accuracy of CNN. L-Softmax does not have a benefit over ST in this case, while EMR still substantially improves the robustness. Note that EMR still needs WD to achieve a satisfactory performance, since the weight decay handles the overall model complexity, while EMR constrains the local weight norms.

\noindent\textbf{Empirical Results for Adversarial Training.}
The experiment with standard training on clean examples suggests that EMR may also have a benefit when adversarial training (AT) is used. Therefore, we compare WD, EMR and L-Softmax using AT \cite{madry2017towards}. In AT, the XE loss has adversarially perturbed input,
\setlength{\belowdisplayskip}{6.3pt} \setlength{\belowdisplayshortskip}{6.3pt}
\setlength{\abovedisplayskip}{0.0pt} \setlength{\abovedisplayshortskip}{0.0pt}
\begin{align}
    \loss = XE(f_{\btheta}(\bx_i+\bdelta_i),y_i), \quad \|\bdelta_i\|_{\infty}\leq\epsilon.
\end{align}
The perturbation is often searched by the PGD attack \cite{madry2017towards}.  Here the same PGD used in the robustness evaluation is adopted in adversarial training for both CNN and MLP.  Fig.~\ref{fig:eff_margin}c-d show the effective margin during AT. 
 The advantage of EMR is decreased when AT is used, but we still observe an increase of effective margin compared with the two baselines.  Tables \ref{tab:mnist_mlp} and \ref{tab:cifar_cnn} (bottom) also show the robustness evaluation for AT. 
  Although the performance gap between EMR and the baseline is decreased compared to ST, we still observe improved robustness for both models using EMR. See the supplemental for the clean and robust test accuracy curve during AT of CNN and MLP.
\section{Effective Margin Regularization for Large-Scale DNNs}
The success of EMR on CNN and MLP provides a motivation to adopt EMR for training large-scale DNNs. One major issue of applying EMR to large-scale models is that the computation of effective weight matrices needs a loop over image categories, which is not scalable to large DNNs \abc{with many classes}. Thus, we propose an approximation of EMR that does not need a loop, which is more amenable to large-scale DNNs. Define $\bl_i$ as the logit  vector for the $i$th sample, and $h_i = \sum_jp_{ij}l_{ij}(\bx_i)$ as the weighted logit mean, where $\sum_j p_{ij}=1$ is a constant weight vector in a ($K-1$)-dim simplex. The gradient of $h_i$ with respect to $\bx_i$ is $\nabla_{\bx} h_i=\sum_{j}p_{ij}\bw_i^{(j)}$, and its squared $l_2$ norm is
\setlength{\belowdisplayskip}{1.8pt} \setlength{\belowdisplayshortskip}{1.8pt}
\setlength{\abovedisplayskip}{1.8pt} \setlength{\abovedisplayshortskip}{1.8pt}
\begin{align}
    \hat\loss_{EMR}(\bx_i)
    =\abc{\|\nabla_{\bx} \sum_jp_{ij}l_{ij}(\bx_i)\|_2^2}
  =\sum\nolimits_{j}\sum\nolimits_{k}p_{ij}p_{ik}\langle\bw_i^{(j)},\bw_i^{(k)} \rangle.
    \label{eqn:appprx_emr}
\end{align}
We take this quantity as an approximation to EMR in (\ref{eqn:emr}), \abc{which implicitly computes the effective weight matrix via the gradient of the logits}. The original EMR regularizes the self-product term in the summation of (\ref{eqn:appprx_emr}), i.e.,
    $\loss_{EMR}(\bx_i)=\sum\nolimits_{j}\langle\bw_i^{(j)},\bw_i^{(j)} \rangle$.
Thus the difference between $\hat\loss_{EMR}$ and $\loss_{EMR}$ is the cross product term between $\bw_i^{(j)}$ and $\bw_i^{(k)}$. Minimizing the cross product is helpful for the classification task because it decreases the cosine \abc{similarity} 
between different categories' weights. 

In (\ref{eqn:appprx_emr}), 
$p_{ij}$ will control the weight for the summation in $\hat\loss_{EMR}$, and \abc{intuitively} higher weights should be applied to the larger logits. Thus, we compute the $p_{ij}$ by a softmax function whose input is $\bl_i$ divided by a temperature $t$. Note that this computation is detached from the gradient computation graph, since we require that $p_{ij}$ is constant. The temperature parameter controls the weight of product terms in $\hat\loss_{EMR}$: if $t\to 0$, we only have the prediction's squared weight norm, i.e., $\hat\loss_{EMR}=\max_{j\in K}\|\bw_i^{(j)}\|_2^2$; if $t\to \infty$, we have a summation of $\langle\bw_i^{(j)},\bw_i^{(k)}\rangle$ for all $i,j$ in $\hat\loss_{EMR}$. In the empirical study, we find that selecting an appropriate temperature parameter is able to improve the performance of EMR. In the supplemental we show the performance of the approximate EMR (Approx-EMR) on CNN and MLP with ST and AT. The approximation achieves a comparable performance in both models and even better robustness in the CNN trained with AT. 

Another crucial problem with EMR is that the ``training mode'' of a DNN with batch normalization will incur correlations among the batch of samples, since the batch normalization is used with the current batch’s mean and variance. To avoid the intertwined gradients, \emph{we use the evaluation mode in the forward propagation when computing EMR}, where we normalize the input batch with the running mean and variance. In this way, we treat the DNN as a locally linear function, and EMR still has the meaning of regularizing local weight norms. Moreover, using evaluation mode in the effective norm computation is faster than using training mode. See the pseudo-code is in the supplemental.
\vspace{-0.3cm}
\section{Large-Scale Experiments}
\vspace{-0.3cm}
We evaluate the proposed EMR on large-scale DNNs and show that EMR improves upon AT, TRADES, two recent baselines \cite{ross2018improving,pang2020boosting} that also aim to optimize the effective margin and strong adversarial defense methods \cite{wang2019improving,liu2021probabilistic}. 
\subsection{Experimental Setting}
\noindent\textbf{Architectures and Datasets.} We evaluate the effectiveness of EMR with ResNet18 \cite{he2016deep} and WideResNet-34-10 \cite{zagoruyko2016wide} following existing work on adversarial robustness \cite{madry2017towards}. The experiment is run on CIFAR10 \cite{krizhevsky2009learning}, consisting of 50k training  and 10k test images. There are 10 object categories in CIFAR10 and each category has 5000 training and 1000 test images.  Both adversarial training with PGD \cite{madry2017towards} and TRADES PGD \cite{zhang2019theoretically} are used in the experiment. 

\noindent\textbf{Training Setting.} We use a standard SGD-Momentum optimizer and XE loss for training. All experiments use an initial learning rate of 0.1 and batch size of 128. For AT-PGD, we train the network for 100 epochs and the learning rate is divided by 10 at epochs 60 and 90. For TRADES, we train the network for 130 epochs and the learning rate is divided by 10 at epochs 60 and 120. 
Note that for EMR, when the learning rate is decayed, $\lambda_{EMR}$ is also divided by 10. To avoid robust overfitting \cite{rice2020overfitting}, we split the original training set into a training and validation set with 48k and 2k images respectively, and select a model with the best robust test accuracy on the validation set. If not mentioned, all methods are evaluated using the model selection based on the validation set for a fair comparison.

\noindent\textbf{Evaluation.} FGSM \cite{goodfellow2014explaining}, PGD \cite{madry2017towards} and AutoAttack \cite{croce2020robustbench} are used to evaluate the adversarial robustness of a DNN. In FGSM, PGD and AutoAttack, we use the $l_{\infty}$ norm as the metric to bound the adversarial perturbation. 
The $l_{\infty}$ norm PGD attack uses gradient ascent to increase the loss by updating the input image and projecting it to an $\epsilon$-bounded $l_{\infty}$ ball.
\CUT{ The $t$-th step of PGD attack is
\begin{align}
    \bdelta_t = \nabla_{\bx}\loss(f_{\btheta}(\bx_t),y),\quad \bx_{t+1} = \text{Proj}_{\mathcal{S}_{\bx}}(\bx_t+\alpha\text{sign}(\bdelta_t)),
\end{align}
where sign is the sign function and $\alpha$ is the step size for the PGD attack.
} 
FGSM is a special case of PGD using only 1 iteration.
AutoAttack is an ensemble of parameter-free attacks consisting of Auto-PGD$_{CE}$, Auto-PGD with Difference of Logits Ratio loss, FAB \cite{croce2020minimally} and Squared Attack \cite{andriushchenko2020square}. In our experiment, we use $\epsilon=8/255=0.031$ and $\alpha=2/255=0.0078$ in FGSM and PGD, and $\epsilon=0.031$ for AutoAttack. We notice that PGD without random start is more effective than PGD with random start so our PGD evaluation starts from the input image without random noise. 

\noindent\textbf{Baselines.} To make a fair comparison between EMR and baseline with WD, we search the $\lambda_{WD}$ from 2e-4 to 1e-3, so that the improvement of EMR upon WD is not  a result of weak WD regularization. Besides AT and TRADES, we compare EMR with Input Gradient Regularization (IGR) \cite{ross2018improving} and Hypersphere Embedding (HE) \cite{pang2020boosting}. IGR regularizes the squared $l_2$ norm of the \emph{loss}, instead of the network output as with EMR, with respect to the input. HE applies a normalization function to the feature, i.e., output of the penultimate layer, and the classification layer's weight, so that the XE loss is only determined by the cosine similarity. We compare the performance of HE and our EMR in the WideResNet experiment using their official implementation. For IGR and HE, we use their default training parameters. For EMR, we select the hyperparameters with grid search. The specific hyperparameter settings are reported in the supplemental.

\CUT{
\begin{table}[t]
    \centering
    \small
    \begin{tabular}{l|c|c|c|c|c}
         & Clean Acc. & FGSM & PGD10 & PGD100 & AutoAttack  \\
         \hline
        AT & 83.39 & 56.95 & 50.88 & 50.07 & 46.90\\
        AT+IGR\cite{ross2018improving} & \textbf{84.01} & 56.97 & 51.03 & 49.66 & 46.52\\
        AT+EWR (ours) & 81.71 & 56.39 & 51.97 & 51.18 & \textbf{47.94}\\
        \hline
        TRADES & \textbf{79.68}  & 57.62& 53.67 & 53.00 & 48.56\\
        TRADES+IGR\cite{ross2018improving} & 78.61 & 57.34 & 53.70 & 53.08 & 48.48\\
        TRADES+EWR (ours) & 79.59 & 57.43 & 53.53 & 53.01 & \textbf{48.86}\\
        \hline
    \end{tabular}
    \caption{Evaluation of adversarial robustness using ResNet18 on CIFAR10.}
    \label{tab:resnet18}
\end{table}
}

\begin{table}[t]
\footnotesize
    \centering
    \begin{tabular}{l|c|c|c|c|c}
         & Clean Acc. & FGSM & PGD10 & PGD100 &AutoAttack  \\
         \hline
        AT & 87.35 & 59.97  & 53.55& 52.30 & 50.31 \\
        AT+IGR\cite{ross2018improving} & \textbf{87.49} & 60.32  & 53.49& 52.42  & 50.42 \\
        AT+HE\cite{pang2020boosting} & 84.53 & 64.07  & 60.36& 59.80 & 51.88 \\
        AT+EMR (ours) & 85.74 & 60.67  & 55.43& 54.62 & \textbf{52.20} \\
        \hline
        TRADES & 82.95 & 60.65  & 56.71& 56.17 & 52.19 \\
        TRADES+IGR\cite{ross2018improving} & \textbf{84.18} & 61.16  & 56.67& 55.90  & 52.41 \\
        TRADES+HE\cite{pang2020boosting} & 79.61 & 60.95  & 58.23& 57.99 & 51.49 \\
        TRADES+EMR (ours) & 83.03 & 60.89  & 57.27& 56.89 & \textbf{52.73} \\
        \hline
        AT+MAIL \cite{liu2021probabilistic} & 86.96 & 60.90 & 55.42 & 54.53 & 45.07 \\
        AT+MAIL+EMR (ours) & \textbf{87.33} & 61.32 & 56.77& 56.00& \textbf{46.25} \\
        \hline
        TRADES+MAIL \cite{liu2021probabilistic} & 84.82 & 60.44 & 55.35 & 54.69 & 51.97 \\
        TRADES+MAIL+EMR (ours) & \textbf{85.37} & 61.67 & 56.82 & 56.21& \textbf{53.29} \\
        \hline
        MART \cite{wang2019improving} & \textbf{83.62} & 61.83 & 57.32 & 56.43 & 51.40 \\
        MART+EMR (ours) & 83.55 & 62.87 & 58.09 & 57.43 & \textbf{52.16} \\
        \hline
    \end{tabular}
    \caption{Evaluation of adversarial robustness using WideResNet-34-10 on CIFAR10. The best result under the strongest attack is emphasized with bold text. }
    \label{tab:wideresnet}
\end{table}

\CUT{
\begin{table}[b]
    \centering
    \vspace{-0.2cm}
    \begin{tabular}{l|c|c|c|c|c}
         & Clean Acc. & FGSM & PGD10 & PGD100  &AA  \\
         \hline
        AT+MAIL & 86.96 & 60.90 & 55.42 & 54.53 & 45.07 \\
        AT+MAIL+EWR (ours) & \textbf{87.33} & 61.32 & 56.77& 56.00& \textbf{46.25} \\
        \hline
        TRADES+MAIL & 84.82 & 60.44 & 55.35 & 54.69 & 51.97 \\
        TRADES+MAIL+EWR (ours) & \textbf{85.37} & 61.67 & 56.82 & 56.21& \textbf{53.29} \\
        \hline
        MART & 83.89 & - & - & - & 52.08 \\
        MART+EMR (ours) & - & - & - & - & - \\
        \hline
    \end{tabular}
    \caption{Result of EMR when combined with  MAIL.}
    \label{tab:mail}
\end{table}
}
\subsection{Experimental Results}
Table \ref{tab:wideresnet} compares the result of vanilla AT, IGR, HE and our EMR using WideResNet-34-10 \cite{madry2017towards}. HE and EMR improve the performance over the vanilla AT, while EMR achieves the best result. For TRADES, HE and IGR do not improve upon the baseline by a large margin, while EMR still achieves the best result. Note that HE has the best robust accuracy under PGD attack, but it is easily attacked by AutoAttack, a stronger and more reliable attack than PGD so it has become a more important evaluation method than PGD for adversarial robustness in recent years \cite{cui2021learnable,liu2021probabilistic}. We also find that HE does not work well with TRADES adversarial training, which is consistent with the claim in their official code repository. See the result of using ResNet18 in the supplemental. 

\begin{figure}
    \centering
    \includegraphics[width=1.0\textwidth]{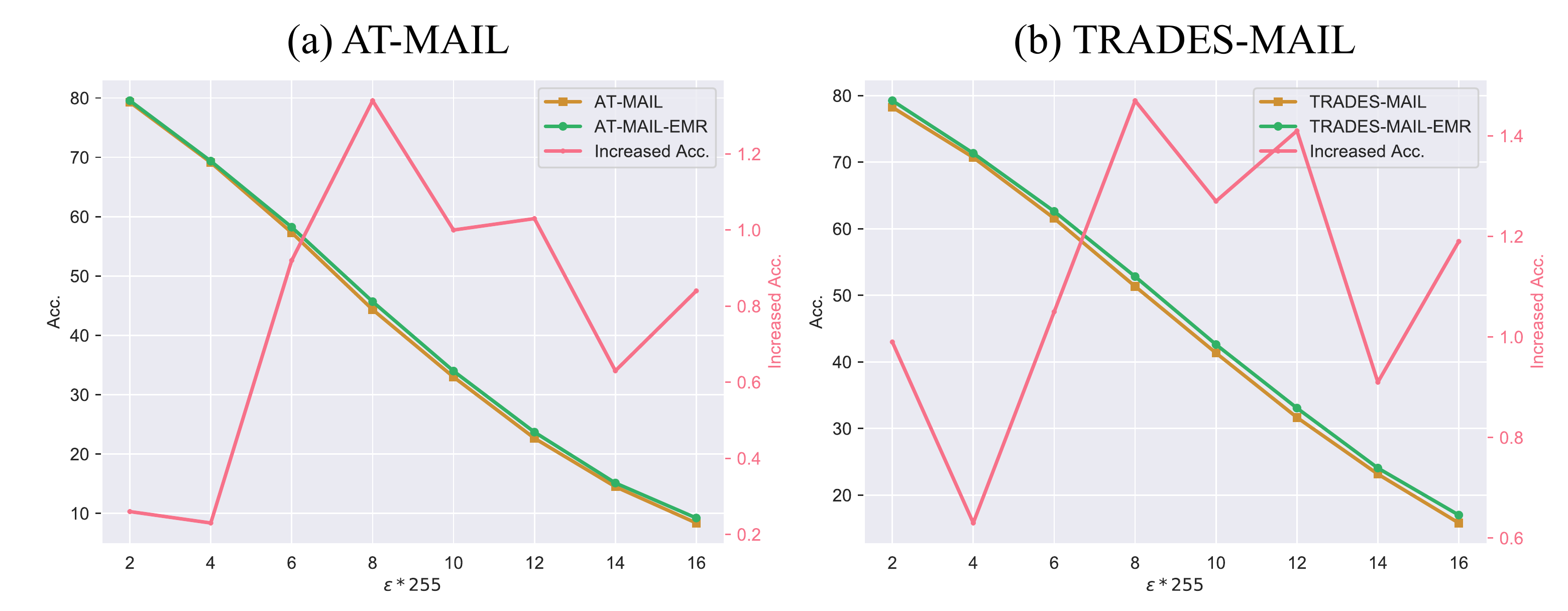}
    \caption{Robust test accuracy of WideResNet under AutoAttack when $\epsilon$ increases. (a) and (b) compare the original MAIL and MAIL+EMR using AT and TRADES, where left y-axis is the absolute accuracy and right y-axis is the relative accuracy of EMR with respect to the MAIL baseline. Our EMR always improves the robustness of MAIL across a large $\epsilon$ range.}
    \label{fig:aa_acc_epsilon}
\end{figure}

Probabilistic margin-aware instance re-weighting learning (MAIL) \cite{liu2021probabilistic} is proposed to weight samples based on the probabilistic margin $p_{iy}-\max_{j\neq y}p_{ij}$, where $p_{ij}$ is the output of softmax function in the XE loss. However, MAIL uses the unnormalized margin in the re-weighting and does not consider the effective margin maximization. Thus, we can apply our EMR in MAIL loss training to control the effective weight norm so that the re-weighting is based on the effective margin instead of the unnormalized margin. We use the default settings in the MAIL loss for MAIL and MAIL+EMR. All training images of CIFAR10 are used for training and the evaluation is done on the model of the final epoch. Table \ref{tab:wideresnet} shows the result of using EMR in MAIL, which demonstrates that for both AT and TRADES, there is a substantial improvement in adversarial robustness. Fig.~\ref{fig:aa_acc_epsilon} shows the robust accuracy under AutoAttack versus attack budget $\epsilon$ and compares MAIL with MAIL-EMR. EMR always improves the performance in this attack range. \ziquan{In addition, we combine EMR with Misclassification Aware adveRsarial Training (MART) \cite{wang2019improving}, to demonstrate the effectiveness of our EMR, shown in Tab.~\ref{tab:wideresnet}, where EMR also substantially improves the robustness under PGD and AutoAttack.}


\begin{table}[t]
    \centering
    \begin{tabular}{l|c|c|c|c|c|c}
         & Model & Clean Acc. & FGSM & PGD10 & PGD100 & AA \\
         \hline
        TRADES+EWR-clean & ResNet18 & \textbf{79.88} & 57.57 & 53.86 & 53.18 & 48.66\\
        TRADES+EWR-adv & ResNet18 & 79.59 & 57.43 & 53.53 & 53.01 & \textbf{48.88}\\
        \hline
        TRADES+EWR-clean & WRN-34-10 & \textbf{83.97} & 61.26 & 56.51 & 55.76 & 52.55\\
        TRADES+EWR-adv & WRN-34-10 & 83.03 & 60.89  & 57.27& 56.89 & \textbf{52.73}\\
        \hline
    \end{tabular}
    \caption{Comparison between TRADES+EMR using clean and adversarial images.}
    \label{tab:trades_clean_adv}
\end{table}
\subsection{Ablation Studies}
\CUT{
\begin{figure}[t]
    \centering
    \includegraphics[width=0.5\textwidth]{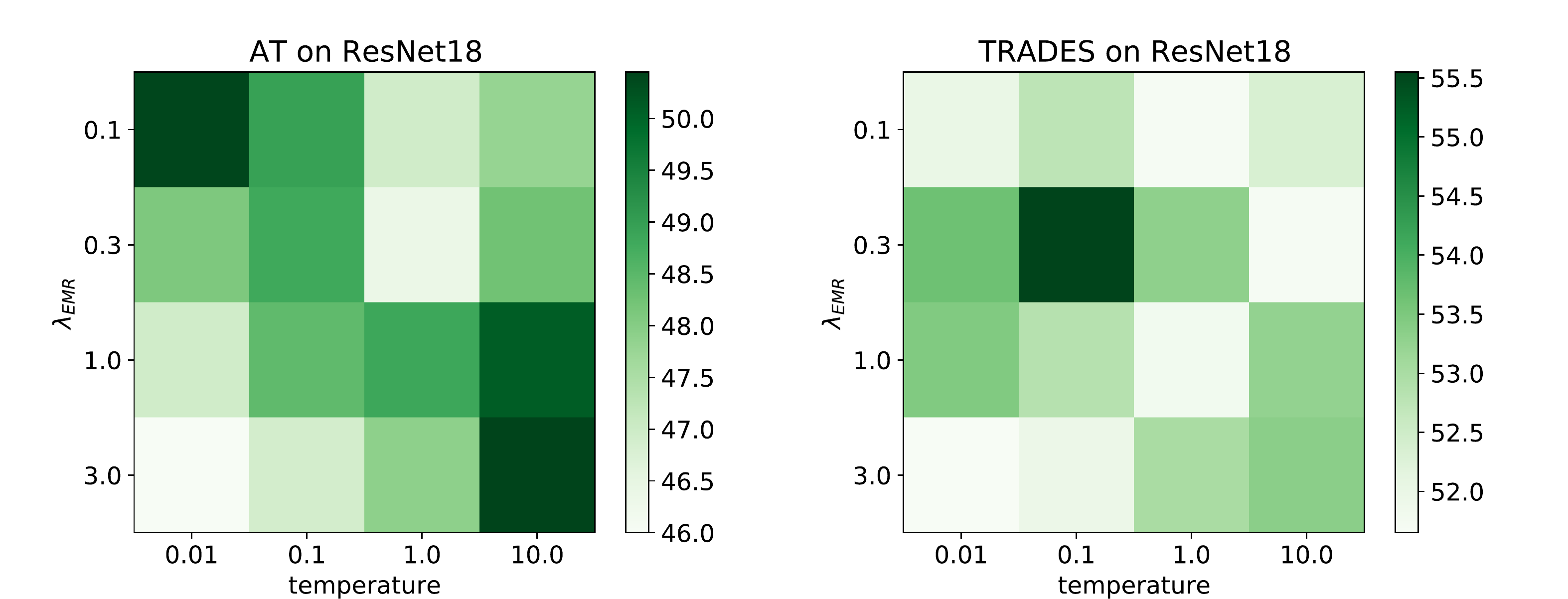}
    \includegraphics[width=0.5\textwidth]{figure/MAIL_epsilon_acc.pdf}
    \caption{Robust test accuracy under PGD10 attack when training a ResNet18 on CIFAR10 with different hyperparameters.}
    \label{fig:hyperpara_emr}
\end{figure}
}

For TRADES+EMR, we study the difference between computing EMR using clean and adversarial examples, since both images are used in TRADES training. Table \ref{tab:trades_clean_adv} shows the results. 
EMR with clean examples has a better clean accuracy, while EMR with adversarial examples has a better robust accuracy. Since our target is the adversarial robustness, we use
latter setting of EMR.
To evaluate the parameter sensitivity of EMR, we evaluate the robustness of ResNet18 when selecting the hyperparameters with grid search. In the supplemental, we show the robust accuracy of different combinations of temperature $t$ and $\lambda_{EMR}$. In AT, selecting a large temperature generally improves the performance. We also report the computational time of EMR combined with IGR and HE in the supplemental.
\section{Conclusion}
This paper investigates the effective margin in training DNNs with the XE loss. Our experiment shows that existing methods do not adequately maximize the effective margin. Therefore, we propose EMR to maximize the effective margin and learn an adversarially robust DNN. On both MLP and CNN, EMR shows a clear strength over WD and L-Softmax in terms of both effective margin and adversarial robustness. On large-scale models, we demonstrate the efficacy of EMR by comparing with 9 strong baselines. We will explore a fast and general EMR in future work so that our method can be applied to larger models.
\section*{Acknowledgement}
This work was supported by a grant from the Research Grants Council of the Hong Kong Special Administrative Region, China (Project No. CityU 11215820).

\clearpage
%
%
\bibliographystyle{splncs04}
\bibliography{egbib}

\begin{thebibliography}{10}
\providecommand{\url}[1]{\texttt{#1}}
\providecommand{\urlprefix}{URL }
\providecommand{\doi}[1]{https://doi.org/#1}

\bibitem{andriushchenko2020square}
Andriushchenko, M., Croce, F., Flammarion, N., Hein, M.: Square attack: a
  query-efficient black-box adversarial attack via random search. In: European
  Conference on Computer Vision. pp. 484--501. Springer (2020)

\bibitem{athalye2018obfuscated}
Athalye, A., Carlini, N., Wagner, D.: Obfuscated gradients give a false sense
  of security: Circumventing defenses to adversarial examples. In:
  International conference on machine learning. pp. 274--283. PMLR (2018)

\bibitem{bachmann2021uniform}
Bachmann, G., Moosavi-Dezfooli, S.M., Hofmann, T.: Uniform convergence,
  adversarial spheres and a simple remedy. In: International Conference on
  Machine Learning. pp. 490--499. PMLR (2021)

\bibitem{buckman2018thermometer}
Buckman, J., Roy, A., Raffel, C., Goodfellow, I.: Thermometer encoding: One hot
  way to resist adversarial examples. In: International Conference on Learning
  Representations (2018)

\bibitem{carlini2017towards}
Carlini, N., Wagner, D.: Towards evaluating the robustness of neural networks.
  In: 2017 ieee symposium on security and privacy (sp). pp. 39--57. IEEE (2017)

\bibitem{chizat2020implicit}
Chizat, L., Bach, F.: Implicit bias of gradient descent for wide two-layer
  neural networks trained with the logistic loss. In: Conference on Learning
  Theory. pp. 1305--1338. PMLR (2020)

\bibitem{crammer2001algorithmic}
Crammer, K., Singer, Y.: On the algorithmic implementation of multiclass
  kernel-based vector machines. Journal of machine learning research
  \textbf{2}(Dec),  265--292 (2001)

\bibitem{croce2020robustbench}
Croce, F., Andriushchenko, M., Sehwag, V., Debenedetti, E., Flammarion, N.,
  Chiang, M., Mittal, P., Hein, M.: Robustbench: a standardized adversarial
  robustness benchmark. arXiv preprint arXiv:2010.09670  (2020)

\bibitem{croce2020minimally}
Croce, F., Hein, M.: Minimally distorted adversarial examples with a fast
  adaptive boundary attack. In: International Conference on Machine Learning.
  pp. 2196--2205. PMLR (2020)

\bibitem{croce2020reliable}
Croce, F., Hein, M.: Reliable evaluation of adversarial robustness with an
  ensemble of diverse parameter-free attacks. In: International conference on
  machine learning. pp. 2206--2216. PMLR (2020)

\bibitem{cui2021learnable}
Cui, J., Liu, S., Wang, L., Jia, J.: Learnable boundary guided adversarial
  training. In: Proceedings of the IEEE/CVF International Conference on
  Computer Vision. pp. 15721--15730 (2021)

\bibitem{dhillon2018stochastic}
Dhillon, G.S., Azizzadenesheli, K., Lipton, Z.C., Bernstein, J., Kossaifi, J.,
  Khanna, A., Anandkumar, A.: Stochastic activation pruning for robust
  adversarial defense. arXiv preprint arXiv:1803.01442  (2018)

\bibitem{Ding2020MaxMarginA}
Ding, G.W., Sharma, Y., Lui, K.Y.C., Huang, R.: Max-margin adversarial (mma)
  training: Direct input space margin maximization through adversarial
  training. ArXiv  \textbf{abs/1812.02637} (2020)

\bibitem{du2018algorithmic}
Du, S.S., Hu, W., Lee, J.D.: Algorithmic regularization in learning deep
  homogeneous models: Layers are automatically balanced. Advances in Neural
  Information Processing Systems  \textbf{31} (2018)

\bibitem{goodfellow2014explaining}
Goodfellow, I.J., Shlens, J., Szegedy, C.: Explaining and harnessing
  adversarial examples. arXiv preprint arXiv:1412.6572  (2014)

\bibitem{gowal2021improving}
Gowal, S., Rebuffi, S.A., Wiles, O., Stimberg, F., Calian, D.A., Mann, T.A.:
  Improving robustness using generated data. Advances in Neural Information
  Processing Systems  \textbf{34} (2021)

\bibitem{guo2018countering}
Guo, C., Rana, M., Cisse, M., van~der Maaten, L.: Countering adversarial images
  using input transformations. In: International Conference on Learning
  Representations (2018)

\bibitem{he2016deep}
He, K., Zhang, X., Ren, S., Sun, J.: Deep residual learning for image
  recognition. In: Proceedings of the IEEE conference on computer vision and
  pattern recognition. pp. 770--778 (2016)

\bibitem{huang2021exploring}
Huang, H., Wang, Y., Erfani, S., Gu, Q., Bailey, J., Ma, X.: Exploring
  architectural ingredients of adversarially robust deep neural networks.
  Advances in Neural Information Processing Systems  \textbf{34} (2021)

\bibitem{ilyas2019adversarial}
Ilyas, A., Santurkar, S., Tsipras, D., Engstrom, L., Tran, B., Madry, A.:
  Adversarial examples are not bugs, they are features. Advances in neural
  information processing systems  \textbf{32} (2019)

\bibitem{krizhevsky2009learning}
Krizhevsky, A., Hinton, G., et~al.: Learning multiple layers of features from
  tiny images  (2009)

\bibitem{krogh1991simple}
Krogh, A., Hertz, J.: A simple weight decay can improve generalization.
  Advances in neural information processing systems  \textbf{4} (1991)

\bibitem{kurakin2016adversarial}
Kurakin, A., Goodfellow, I., Bengio, S.: Adversarial machine learning at scale.
  arXiv preprint arXiv:1611.01236  (2016)

\bibitem{lecun1998mnist}
LeCun, Y.: The mnist database of handwritten digits. http://yann. lecun.
  com/exdb/mnist/  (1998)

\bibitem{li2021towards}
Li, Y., Min, M.R., Lee, T., Yu, W., Kruus, E., Wang, W., Hsieh, C.J.: Towards
  robustness of deep neural networks via regularization. In: Proceedings of the
  IEEE/CVF International Conference on Computer Vision. pp. 7496--7505 (2021)

\bibitem{liu2021probabilistic}
Liu, F., Han, B., Liu, T., Gong, C., Niu, G., Zhou, M., Sugiyama, M., et~al.:
  Probabilistic margins for instance reweighting in adversarial training.
  Advances in Neural Information Processing Systems  \textbf{34} (2021)

\bibitem{liu2017sphereface}
Liu, W., Wen, Y., Yu, Z., Li, M., Raj, B., Song, L.: Sphereface: Deep
  hypersphere embedding for face recognition. In: Proceedings of the IEEE
  conference on computer vision and pattern recognition. pp. 212--220 (2017)

\bibitem{liu2016large}
Liu, W., Wen, Y., Yu, Z., Yang, M.: Large-margin softmax loss for convolutional
  neural networks. In: ICML. vol.~2, p.~7 (2016)

\bibitem{liu2020improve}
Liu, Z., Cui, Y., Chan, A.B.: Improve generalization and robustness of neural
  networks via weight scale shifting invariant regularizations. arXiv preprint
  arXiv:2008.02965  (2020)

\bibitem{lu2017safetynet}
Lu, J., Issaranon, T., Forsyth, D.: Safetynet: Detecting and rejecting
  adversarial examples robustly. In: Proceedings of the IEEE international
  conference on computer vision. pp. 446--454 (2017)

\bibitem{lyu2019gradient}
Lyu, K., Li, J.: Gradient descent maximizes the margin of homogeneous neural
  networks. arXiv preprint arXiv:1906.05890  (2019)

\bibitem{ma2018characterizing}
Ma, X., Li, B., Wang, Y., Erfani, S.M., Wijewickrema, S., Schoenebeck, G.,
  Song, D., Houle, M.E., Bailey, J.: Characterizing adversarial subspaces using
  local intrinsic dimensionality. arXiv preprint arXiv:1801.02613  (2018)

\bibitem{madry2017towards}
Madry, A., Makelov, A., Schmidt, L., Tsipras, D., Vladu, A.: Towards deep
  learning models resistant to adversarial attacks. arXiv preprint
  arXiv:1706.06083  (2017)

\bibitem{moosavi2016deepfool}
Moosavi-Dezfooli, S.M., Fawzi, A., Frossard, P.: Deepfool: a simple and
  accurate method to fool deep neural networks. In: Proceedings of the IEEE
  conference on computer vision and pattern recognition. pp. 2574--2582 (2016)

\bibitem{pang2018towards}
Pang, T., Du, C., Dong, Y., Zhu, J.: Towards robust detection of adversarial
  examples. Advances in Neural Information Processing Systems  \textbf{31}
  (2018)

\bibitem{pang2019rethinking}
Pang, T., Xu, K., Dong, Y., Du, C., Chen, N., Zhu, J.: Rethinking softmax
  cross-entropy loss for adversarial robustness. arXiv preprint
  arXiv:1905.10626  (2019)

\bibitem{pang2020boosting}
Pang, T., Yang, X., Dong, Y., Xu, K., Zhu, J., Su, H.: Boosting adversarial
  training with hypersphere embedding. Advances in Neural Information
  Processing Systems  \textbf{33},  7779--7792 (2020)

\bibitem{qin2019adversarial}
Qin, C., Martens, J., Gowal, S., Krishnan, D., Dvijotham, K., Fawzi, A., De,
  S., Stanforth, R., Kohli, P.: Adversarial robustness through local
  linearization. Advances in Neural Information Processing Systems  \textbf{32}
  (2019)

\bibitem{rebuffi2021data}
Rebuffi, S.A., Gowal, S., Calian, D.A., Stimberg, F., Wiles, O., Mann, T.A.:
  Data augmentation can improve robustness. Advances in Neural Information
  Processing Systems  \textbf{34} (2021)

\bibitem{rice2020overfitting}
Rice, L., Wong, E., Kolter, Z.: Overfitting in adversarially robust deep
  learning. In: International Conference on Machine Learning. pp. 8093--8104.
  PMLR (2020)

\bibitem{ross2018improving}
Ross, A., Doshi-Velez, F.: Improving the adversarial robustness and
  interpretability of deep neural networks by regularizing their input
  gradients. In: Proceedings of the AAAI Conference on Artificial Intelligence.
  vol.~32 (2018)

\bibitem{roth2019odds}
Roth, K., Kilcher, Y., Hofmann, T.: The odds are odd: A statistical test for
  detecting adversarial examples. In: International Conference on Machine
  Learning. pp. 5498--5507. PMLR (2019)

\bibitem{samangouei2018defense}
Samangouei, P., Kabkab, M., Chellappa, R.: Defense-gan: Protecting classifiers
  against adversarial attacks using generative models. arXiv preprint
  arXiv:1805.06605  (2018)

\bibitem{song2017pixeldefend}
Song, Y., Kim, T., Nowozin, S., Ermon, S., Kushman, N.: Pixeldefend: Leveraging
  generative models to understand and defend against adversarial examples.
  arXiv preprint arXiv:1710.10766  (2017)

\bibitem{szegedy2013intriguing}
Szegedy, C., Zaremba, W., Sutskever, I., Bruna, J., Erhan, D., Goodfellow, I.,
  Fergus, R.: Intriguing properties of neural networks. arXiv preprint
  arXiv:1312.6199  (2013)

\bibitem{tsipras2018robustness}
Tsipras, D., Santurkar, S., Engstrom, L., Turner, A., Madry, A.: Robustness may
  be at odds with accuracy. arXiv preprint arXiv:1805.12152  (2018)

\bibitem{wang2018additive}
Wang, F., Cheng, J., Liu, W., Liu, H.: Additive margin softmax for face
  verification. IEEE Signal Processing Letters  \textbf{25}(7),  926--930
  (2018)

\bibitem{wang2019improving}
Wang, Y., Zou, D., Yi, J., Bailey, J., Ma, X., Gu, Q.: Improving adversarial
  robustness requires revisiting misclassified examples. In: International
  Conference on Learning Representations (2019)

\bibitem{xie2020smooth}
Xie, C., Tan, M., Gong, B., Yuille, A., Le, Q.V.: Smooth adversarial training.
  arXiv preprint arXiv:2006.14536  (2020)

\bibitem{xie2017mitigating}
Xie, C., Wang, J., Zhang, Z., Ren, Z., Yuille, A.: Mitigating adversarial
  effects through randomization. arXiv preprint arXiv:1711.01991  (2017)

\bibitem{xu2017feature}
Xu, W., Evans, D., Qi, Y.: Feature squeezing: Detecting adversarial examples in
  deep neural networks. arXiv preprint arXiv:1704.01155  (2017)

\bibitem{zagoruyko2016wide}
Zagoruyko, S., Komodakis, N.: Wide residual networks. arXiv preprint
  arXiv:1605.07146  (2016)

\bibitem{zhang2019theoretically}
Zhang, H., Yu, Y., Jiao, J., Xing, E., El~Ghaoui, L., Jordan, M.: Theoretically
  principled trade-off between robustness and accuracy. In: International
  conference on machine learning. pp. 7472--7482. PMLR (2019)

\end{thebibliography}

\section*{Appendix}

\subsection*{A. Experimental Settings}
\paragraph{Model Architecture.} Section 3 uses an MLP and CNN to demonstrate the effective margin maximization effect of EMR. The MLP has 4 hidden layers with 1024 hidden neurons and one output layer. The CNN has an architecture as in Table \ref{tab:cnn_arch}, where the parameter in the convolution layer means CONV(kernel\_size, output\_channel, stride, padding) and in the average pooling layer means Average\_Pooling(kernel\_size, stride). For both models we use ReLU as the activation function.

\begin{figure}[b]
    \centering
    \includegraphics[width=1.0\textwidth]{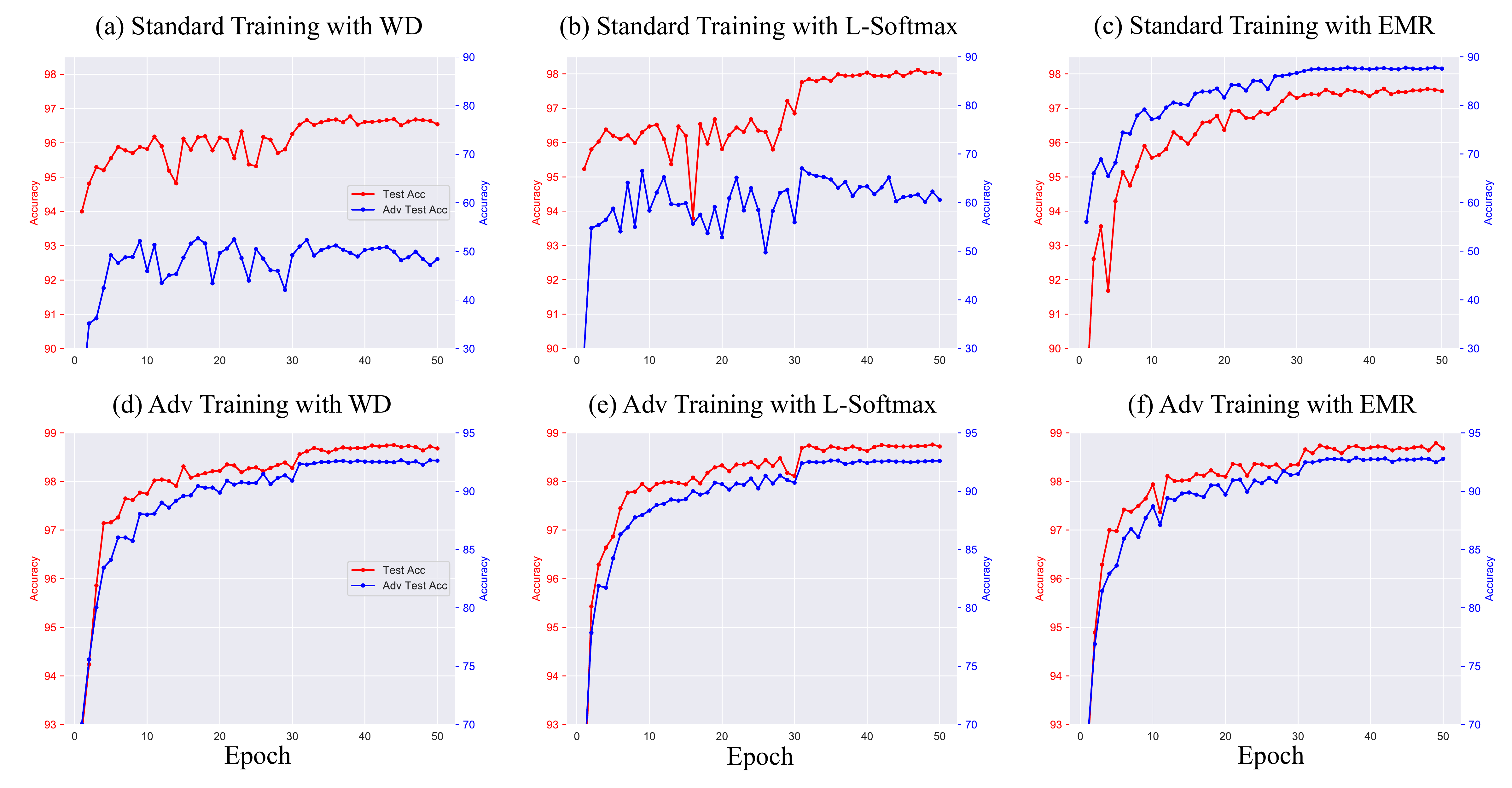}
    \caption{Training curve on MNIST with 4-hidden layer MLP. In normal training, EMR achieves significantly higher robust test accuracy than two baselines and is even comparable with that of adversarial training.}
    \label{fig:mnist_train_curve}
\end{figure}

\begin{figure}[t]
    \centering
    \includegraphics[width=1.0\textwidth]{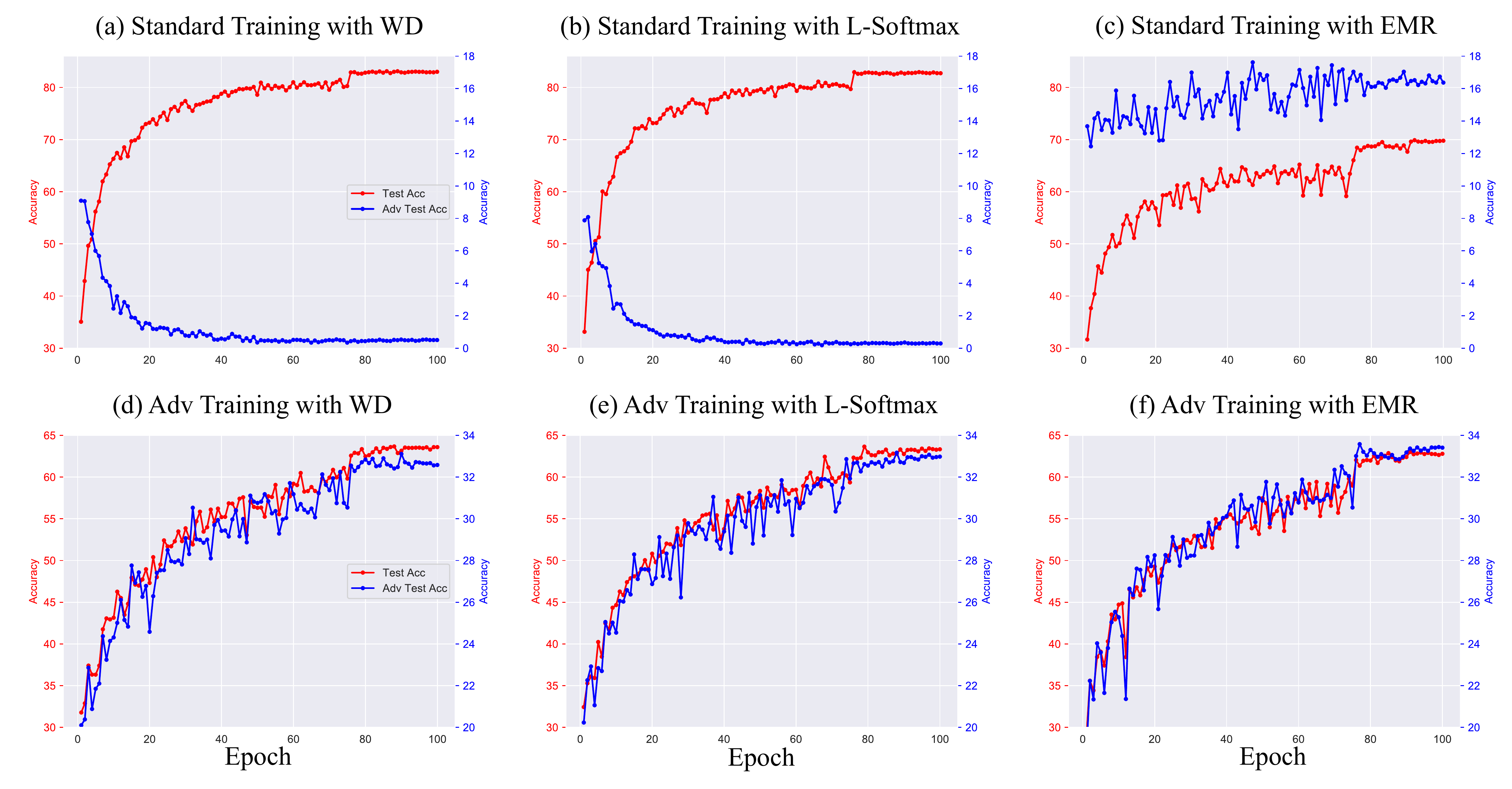}
    \caption{Training curve on CIFAR10 with 4-hidden layer CNN. In standard training, EMR has 
    a higher robust test accuracy than training with WD or L-Softmax. In adversarial training, EMR also achieves the best robust accuracy.
    }
    \label{fig:cifar_train_curve}
\end{figure}

\begin{table}[h]
    \centering
    \begin{tabular}{c|c}
    \hline
         Layer 1 & CONV(5,32,1,padding=None) \\
         \hline
         Layer 2 & CONV(5,64,1,padding=None) \\
         \hline
         Layer 3 & Average\_Pooling(2,2,padding=None) \\
         \hline
         Layer 4 & CONV(3,128,1,padding=None) \\
         \hline
         Layer 5 & CONV(3,128,1,padding=None) \\
         \hline
         Layer 6 & Global\_Average\_Pooling \\
         \hline
         Layer 6 & Linear(128,10) \\
         \hline
    \end{tabular}
    \caption{The architecture of CNN in Section 3. }
    \label{tab:cnn_arch}
\end{table}
\begin{algorithm}[b]
 \caption{Adv. Training with Effective Margin Regularization}
  \KwIn{Training data $\datadis_{tr}$, EMR parameter $\lambda_{EMR}$, EMR temperature $t$, learning rate $\eta$, beta of TRADES $\beta$ }
  \KwOut{Model parameters $\btheta$}
  Initialize model parameters;
  \\
    \For{$i=1,\dots,N_e$}
    {
      \textcolor{red}{Adjust $\eta$ and $\lambda_{EMR}$};\\
      Split $\datadis_{tr}$ into $N_B=$ceil$(N_{tr}/B)$ batches;
      \\
      \For{$b=1,\dots,N_B$}
      {
        Generate adversarial examples $\{\tilde\bx_i,y_i\}_{i=1}^{B}$;\\
        \If{\textcolor{orange}{AT}} {\textcolor{orange}{$\loss=\frac{1}{B}\sum_{i=1}^B XE(f_{\btheta}(\tilde\bx_i),y_i)$;}}
        \If{\textcolor{blue}{TRADES}} {\textcolor{blue}{$\loss=\frac{1}{B}\sum_iXE(f_{\btheta}(\bx_i),y_i)+\beta\frac{1}{B}\sum_i\datadis_{KL}(f_{\btheta}(\tilde\bx_i),f_{\btheta}(\bx_i))$;}}
        \% \textcolor{red}{EMR:}\\
        \textcolor{red}{$\bp_{i}$=softmax($f_{\btheta}(\tilde\bx_i)/t$) and detach the gradient;}\\
        \textcolor{red}{$\loss_{EMR}=\frac{1}{B}\sum_i^B \|\nabla_{\bx}\sum_{j=1}^Kp_{ij}l_{ij}(\tilde\bx_i)\|_2^2$ (\textbf{eval mode});}\\
        \textcolor{red}{$\btheta := \btheta - \eta\nabla_{\btheta}(\loss+\lambda_{EMR}\loss_{EMR})$}
      }
    }
    \label{alg:at_emr}
\end{algorithm}
\paragraph{Hyperparameters. } We searched the hyperparameters for the adversarial training baselines and our method, Table \ref{tab:resnet18_hyper} and \ref{tab:wideresnet_hyper} show the hyperparameters used in the experiment we report. In MAIL, we use the WideResNet-34-10 and default settings in the MAIL loss \cite{liu2021probabilistic}. AT-MAIL has a slope of 30 and bias of 0.07 . TRADES-MAIL has a slope of 5 and bias of 0.05, where $\beta$=5.0. All MAIL experiment uses a weight decay of 2e-4. In AT-MAIL-ERM, $\lambda_{EMR}$=1.0, $t$=1.0. In TRADES-MAIL-ERM, $\lambda_{EMR}$=3.0, $t$=1.0. We use PGD10 attack with a step size of 0.00784 during adversarial training of MAIL and the initial learning rate 0.1 is decayed by 10 at 75 and 90 epoch, with a maximum epoch of 100. In MART+EMR \cite{wang2019improving}, we use the default setting of MART loss from the official code, where $\beta=6.0$, the step size of PGD attack is 0.007 and step is 10. With the MART loss, the WideResNet-34-10 is trained for 90 epochs and the initial learning rate 0.1 is divided by 10 at 75 and 90 epoch, and we let $\lambda_{EMR}=t=$1.0. In LBGAT+EMR \cite{cui2021learnable}, we use the default setting in the official code, where the teacher model is a randomly initialized ResNet18, the student model is a WideResNet34-10, $\beta=$0.0 (without TRADES) and step size of PGD10 is 0.003. We train the model for 100 epochs and divide the initial learning rate 0.1 by 10 at 76 and 91 epoch, and let $\lambda_{EMR}=0.5$ and $t=$1. For the MART and LBGAT baseline, we evaluate the official models released by the authors. Note that we use the $l_{\infty}$-bound adversarial examples with an $\epsilon=0.031$ in training and evaluation of all experiment. The algorithm for adversarial training with EMR is shown in Algorithm~\ref{alg:at_emr}. We include the code to use our method in the supplemental. 

\subsection*{B. More Experiment Result}
Fig.~\ref{fig:cifar_train_curve} and Fig.~\ref{fig:mnist_train_curve} show the clean and robust test accuracy curves during CNN and MLP training. For the MLP, EMR achieves a high robust accuracy at an early stage of training. For the CNN, EMR shows an oscillation of robust test accuracy at an early stage, but the accuracy becomes stable when the learning rate is decayed.

Table \ref{tab:approx_EMR} shows the performance of the approximate EMR (Approx-EMR) on CNN and MLP with ST and AT. The approximation achieves a comparable performance in both models and even better robustness in the CNN trained with AT. Table \ref{tab:resnet18} shows the result of vanilla AT, IGR and our EMR on ResNet18. Compared with AT and IGR, the adversarial robustness (as measured by AutoAttack) is substantially improved when EMR is used. For TRADES, the improvement is not as significant as with AT. Note that IGR does not improve the performance substantially for either AT or TRADES. 

To evaluate the parameter sensitivity of EMR, we evaluate the robustness of ResNet18 when selecting the hyperparameters with grid search. Fig.~\ref{fig:hyperpara_emr} shows the robust accuracy of different combinations of temperature $t$ and $\lambda_{EMR}$. In AT, selecting a large temperature generally improves the performance.

Finally, we evaluate the computational time of EMR. Table \ref{tab:time_compare} shows a comparison between training time of one SGD update of EMR, IGR, HE and the vanilla AT/TRADES when a WideResNet is used. EMR and IGR have approximately the same computational time that is longer than HE, since the loss requires an extra backpropagation. HE has a benefit in the computational time during training, but the normalization layers bring extra computation for the \emph{inference} stage. In contrast, EMR does not need more computation during inference. 

\begin{table}[]
    \centering
    \begin{tabular}{c|c}
         & Hyperparameters \\
         \hline
        AT  & $\lambda_{WD}$=1e-3\\
        AT+IGR & $\lambda_{WD}$=1e-3, $\lambda_{IGR}$=1.0\\
        AT+EWR & $\lambda_{WD}$=5e-4, $\lambda_{EMR}$=0.1, $t$=40.0\\
        \hline
        TRADES  & $\beta$=12.0, $\lambda_{WD}$=5e-4\\
        TRADES+IGR & $\beta$=12.0, $\lambda_{WD}$=5e-4, $\lambda_{IGR}$=1.0 \\
        TRADES+EWR & $\beta$=12.0, $\lambda_{WD}$=5e-4, $\lambda_{EMR}$=0.3, $t$=0.1 \\
        \hline
    \end{tabular}
    \caption{Hyperparameters of ResNet18.}
    \label{tab:resnet18_hyper}
\end{table}

\begin{table}[]
    \centering
    \begin{tabular}{c|c}
         & Hyperparameters \\
         \hline
        AT  & $\lambda_{WD}$=1e-3\\
        AT+IGR & $\lambda_{WD}$=1e-3, $\lambda_{IGR}$=1.0\\
        AT+EWR & $\lambda_{WD}$=5e-4, $\lambda_{EMR}$=1.0, $t$=40.0\\
        \hline
        TRADES  & $\beta$=12.0, $\lambda_{WD}$=5e-4\\
        TRADES+IGR & $\beta$=12.0, $\lambda_{WD}$=5e-4, $\lambda_{IGR}$=1.0\\
        TRADES+EWR & $\beta$=12.0, $\lambda_{WD}$=5e-4, $\lambda_{EMR}$=0.3, $t$=1.0\\
        \hline
    \end{tabular}
    \caption{Hyperparameters of WideResNet.}
    \label{tab:wideresnet_hyper}
\end{table}

\begin{table}[t]
    \centering
    \small
    \begin{tabular}{c|lc|c|c|c|c}
         Model &Training & $\lambda_{WD}$ & Clean Acc. & PGD & $\tilde m_{train}$ &  $\tilde m_{test}$  \\
         \hline
        \multirow{4}{*}{MLP} & ST+EMR$_{0.1}$ & 0.001 & 97.50 & \textbf{87.56} & 4.41$\pm$1.24  & 2.24$\pm$0.98\\
        &ST+Approx-EMR$_{1.0}$ & 0.001 & \textbf{98.44} & 77.09 & 0.85$\pm$0.71  & 1.73$\pm$0.73\\
        \cline{2-7}
        &AT+EMR$_{0.0003}$ & 0.001 & 98.68 & \textbf{92.78} & 3.83$\pm$1.27  & 2.42$\pm$0.99\\
        &AT+Approx-EMR$_{0.0003}$ & 0.001 & \textbf{98.75} & 92.76 & 4.00$\pm$1.30  & 2.35$\pm$0.94\\
        \hline
        \multirow{4}{*}{CNN} & ST+EMR$_{0.01}$&0.0005 & 69.78 & \textbf{16.37} & 0.66$\pm$0.48 & 0.70$\pm$0.50\\
        &ST+Approx-EMR$_{30.0}$&0.0005 & \textbf{71.09} & 15.05 & 0.65$\pm$0.48 & 0.68$\pm$0.50\\
        \cline{2-7}
        &AT+EMR$_{0.001}$ &0.0005& 62.79 & 33.41  & 0.74$\pm$0.64 & 1.08$\pm$0.83\\
        &AT+Approx-EMR$_{0.0003}$ &0.0005& \textbf{63.15} & \textbf{33.63}  & 0.73$\pm$0.62 & 1.07$\pm$0.81\\
        \hline
    \end{tabular}
    \caption{Comparison between EMR and its large-scale approximation.}
    \label{tab:approx_EMR}
\end{table}

\begin{table}[t]
    \centering
    \begin{tabular}{l|c|c|c|c|c}
         & Clean Acc. & FGSM & PGD10 & PGD100 & AutoAttack  \\
         \hline
        AT & 83.39 & 56.95 & 50.88 & 50.07 & 46.90\\
        AT+IGR\cite{ross2018improving} & \textbf{84.01} & 56.97 & 51.03 & 49.66 & 46.52\\
        AT+EWR (ours) & 81.71 & 56.39 & 51.97 & 51.18 & \textbf{47.94}\\
        \hline
        TRADES & \textbf{79.68}  & 57.62& 53.67 & 53.00 & 48.56\\
        TRADES+IGR\cite{ross2018improving} & 78.61 & 57.34 & 53.70 & 53.08 & 48.48\\
        TRADES+EWR (ours) & 79.59 & 57.43 & 53.53 & 53.01 & \textbf{48.86}\\
        \hline
    \end{tabular}
    \caption{Evaluation of adversarial robustness using ResNet18 on CIFAR10.}
    \label{tab:resnet18}
\end{table}

\begin{figure}[t]
    \centering
    \includegraphics[width=0.8\textwidth]{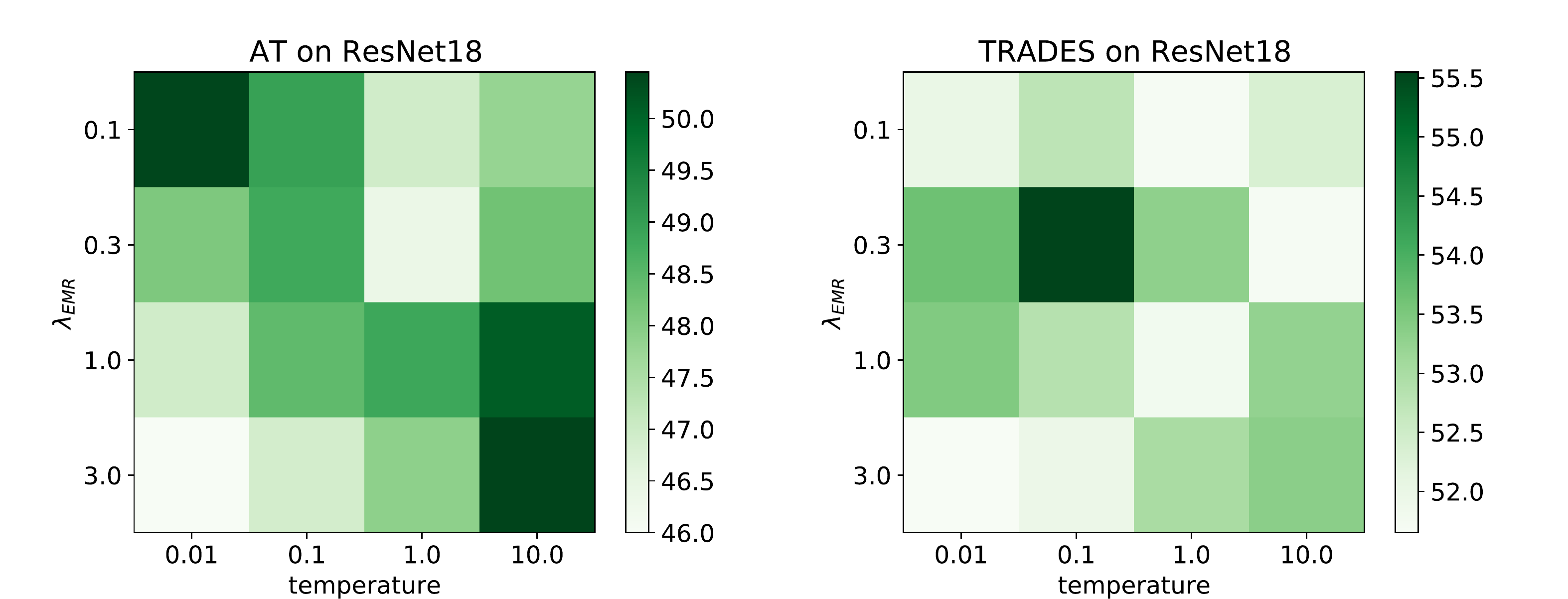}
    \vspace{-0.3cm}
    \caption{Robust test accuracy under PGD10 attack when training a ResNet18 on CIFAR10 with different hyperparameters.}
    \vspace{-0.2cm}
    \label{fig:hyperpara_emr}
\end{figure}

\begin{table}[t]
    \centering
    \begin{tabular}{c|c|c|c|c}
         & Standard & IGR\cite{ross2018improving} & HE\cite{pang2020boosting} & EMR(ours)   \\
         \hline
        AT  & 2.46(.01) & 3.21(.01) & 2.75(.01) & 3.22(.01) \\
        TRADES  & 2.93(.01) & 3.67(.01) & 3.32(.01) & 3.67(.01) \\
        \hline
    \end{tabular}
    \caption{Running time (in second) of one SGD iteration. The time is recorded with WideResNet-34-10 on a Nvidia-V100 with a batch size of 128.}
    \label{tab:time_compare}
\end{table}

\end{document}